\documentclass{article}
\usepackage{graphicx} 
\usepackage{multirow}
\usepackage{amsmath}
\usepackage{authblk}

\usepackage{amssymb}
\usepackage{mathtools}
\usepackage{amsthm}
\title{Forward-Forward Algorithm for Hyperspectral Image Classification: A Preliminary Study}
\author[]{Sidike Paheding }
\author[]{Abel A. Reyes-Angulo }
\affil[]{\textit{Dept. of Applied Computing, Michigan Technological University, Houghton, MI 49931, USA}}

\date{}

\begin{document}

\maketitle

\begin{abstract}
    
    The back-propagation algorithm has long been the de-facto standard in optimizing weights and biases in neural networks, particularly in cutting-edge deep learning models. Its widespread adoption in fields like natural language processing, computer vision, and remote sensing has revolutionized automation in various tasks. The popularity of back-propagation stems from its ability to achieve outstanding performance in tasks such as classification, detection, and segmentation. Nevertheless, back-propagation is not without its limitations, encompassing sensitivity to initial conditions, vanishing gradients, overfitting, and computational complexity. The recent introduction of a forward-forward algorithm (FFA), which computes local goodness functions to optimize network parameters, alleviates the dependence on substantial computational resources and the constant need for architectural scaling. This study investigates the application of FFA for hyperspectral image classification. Experimental results and comparative analysis are provided with the use of the traditional back-propagation algorithm. Preliminary results show the potential behind FFA and its promises.  
\end{abstract}

\section{Introduction}
Deep Learning(DL) \cite{lecun2015deep} has been revolutionizing many different fields due to its ability to achieve unprecedented performance when applied to real-world problems, including applications in agriculture \cite{kamilaris2018deep, maimaitijiang2020soybean}, medicine \cite{ching2018opportunities}, cyber-security \cite{mahdavifar2019application}, and many others \cite{bourilkov2019machine, khan2021trends, siddique2021u, najafabadi2015deep, alom2019state}.

Hyperspectral image (HSI) contains an extensive array of continuous spectral information across numerous narrow bands. The inherent high-dimensionality of hyperspectral data poses substantial obstacles to accurate classification, owing to intricate spectral variations and a scarcity of labeled samples. Deep learning models, specifically convolutional neural networks (CNNs) \cite{lecun1998gradient}, have exhibited remarkable accomplishments in numerous computer vision tasks, notably including the classification of hyperspectral images \cite{hu2015deep, li2019deep}. Nevertheless, the conventional backpropagation algorithm, which is commonly employed for training deep learning models, may confront certain limitations within this domain, such as computational or energy cost, and sensitivity to initial conditions \cite{sexton2000comparative, hinton2022forward}.

By propagating errors in a backward manner through the network, the backpropagation algorithm calculates gradients that guide the adjustment of the model's parameters to minimize the objective function \cite{rumelhart1986learning}. While backpropagation has proven effective in deep learning applications, it encounters difficulties when dealing with hyperspectral data such as limited availability of labeled samples. As a result, there is a need for alternative training approaches that can enhance the performance of deep learning models specifically in the context of hyperspectral image classification.

In this study, we investigate the performance of the forward-forward algorithm (FFA) \cite{hinton2022forward} for the task of hyperspectral image classification. The FFA explores the relationships among input data samples by feeding forward both the original data (positive data) and an alternative version of this data (negative data), encouraging the model to learn robust features that capture the underlying characteristics of the HSI. 

Our initial experiments show that solely using the FFA does not yield better results compared to the backpropagation algorithm. However, considering the advantages of both methods, we propose to combine the forward-forward pass algorithm with traditional backpropagation for hyperspectral image classification. The idea is to incorporate the forward-forward algorithm as an initial learning stage, allowing the model to learn more discriminative features. Subsequently, the model is fine-tuned using the backpropagation algorithm, which refines the learned representations and optimizes the classification performance. To the best of our knowledge, this work is the first attempt to utilize FFA for HSI classification task.

The remaining sections of this paper are structured as follows: Section \ref{sec:ffn} presents an overview of the FFA explored in this work. Section \ref{sec: ffnForHSI} details the proposed hybrid approach. Section \ref{sec: Datasets} presents the benchmark dataset utilized for the experiments. Section \ref{sec: results} provides the experimental setup and analyzes the obtained experimental results. Finally, Section \ref{sec: conclusion} provides a summary of the findings from our study.

\section{Methods}\label{sec:ffn}

\subsection{The backpropagation}
Introduced by Rumelhart et al. \cite{rumelhart1986learning, rumelhart1985learning}, in the backpropagation algorithm for training artificial neural networks, the process involves computing the gradient of the cost function with respect to the network's parameters and using this information to update the weights via gradient descent. Backpropagation has demonstrated strong generalization capabilities and effectiveness in handling non-linearities, making it applicable to a wide range of neural network types including feedforward networks, recurrent networks, and convolutional networks.

\subsection{The Forward-Forward algorithm}
Forward-forward algorithm (FFA) \cite{hinton2022forward} involves substituting the conventional forward and backward passes from the backpropagation algorithm, with two forward passes that function in a parallel manner but on distinct data with opposing objectives. The affirmative pass involves real data and modifies the weights to improve the goodness in each hidden layer, while the negative pass operates on ``negative data'' and modifies the weights to diminish the goodness within every hidden layer. In \cite{hinton2022forward}, two distinct criteria were investigated for measuring quality: the sum of the squared neural activities and the negative sum of the squared activities, although numerous other criteria can also be utilized. In the original FFA, the sum of the squares of the activities in the layer is expressed as
\begin{equation}
    G=\sum_j z^2_{j}, 
\end{equation}

\noindent where $z_i$ represents the   activity of the $j^{th}$ hidden unit. 

Furthermore, the positive and negative passes adjust the weights locally, and the probability of the outputs are expressed as follows:

\begin{equation}
    prob(\text{positive})=\sigma(G-\theta)
\end{equation}

\noindent where $\sigma$ despite a logistic distribution function, and and $\theta$ a given threshold.

\begin{figure}[ht]%
\centering
\includegraphics[width=1\textwidth]{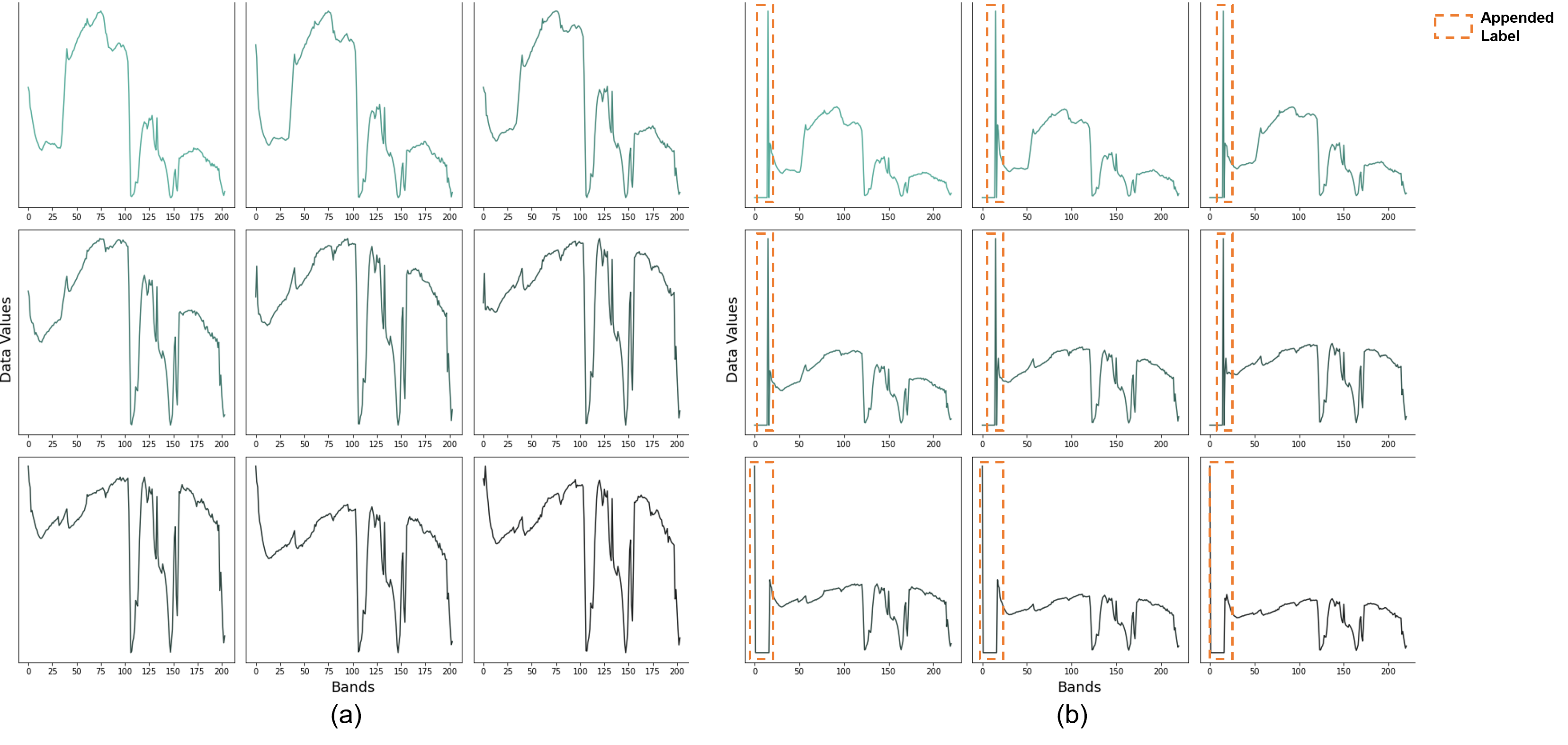}
\caption{Illustration of label embedding method. (a) shows spectral signatures from the Salinas hyperspectral dataset. (b) shows the same samples with the appended label information, which is appended at the beginning of spectral signatures.}\label{fig1}
\end{figure}

To facilitate the contrast of positive and negative data during the supervised training process of FFA, we need to develop a method for merging the data with their corresponding labels. In \cite{hinton2022forward}, Hinton proposed to overlay the label information onto the data itself or embed the label within the input data. However, in this work, we take a different approach by appending the label at one end of the spectral signature of each sample. Moreover, we explore various methods for encoding the label information. For instance, we experiment with one-hot encoding representation, binary representation, and decimal representation. Through the analysis of experimental results, it is found that using the one-hot encoding representation to append the label information with the hyperspectral signature data yields better outcomes. As a result, all the experiments reported in this work utilizing FFA are trained using this approach.
Figure \ref{fig1} visually depicts the imputation method employed, where the label-encoded information is appended at one end of the pixel's hyperspectral signatures.

\section{FFA for hyperspectral image classification}\label{sec: ffnForHSI}
In this preliminary study, we contemplate the implementation of FFA with the use of fully connected layers and 1D convolutional layers for HSI classification.
The fully connected layer performs a linear transformation on the input data, usually in a vector shape, through weights matrix multiplication operations, in addition to a bias term. The fully connected layers comprise the contributions of all inputs for a final prediction. 
In contrast, convolutional layers in CNN can be used to capture local dependencies in sequential data, such as time series or text. Unlike fully connected layers that operate on the entire input, CNN considers the local receptive field of the input at a time. This allows them to extract features that are sensitive to local patterns and variations.

\subsection{Fully Connected FFA network for pixel-wise HSI classification}
The classification of hyperspectral images poses a substantial challenge attributed to the data's high dimensionality and spectral complexity. Deep learning architectures have shown promise in extracting discriminative features from hyperspectral images. Nonetheless, the efficacy of these architectures is heavily contingent upon the quality of the learned representations. In this study, we explore the use of the FFA with Fully Connected layers to enhance for HSI classification task. FFA comprises the use of a few hidden layers to extract features from the HSI data, scale it to a latent space, and produce the final pixel-wise classification. A total of 3 hidden layers were used in our FFA, with the following number of units: 784, 500, and 500, respectively.

\subsection{Convolutional FFA network for pixel-wise HSI classification}
In this work, we explore the implementation of the aforementioned types of neural network layers, limited to 1D data. Convolutional layers for 1D data are usually implemented as neural networks that convolve the input of the hyperspectral image with a set of learnable filters. These layers detect local relationships and patterns within the interaction of the hyperspectral bands per pixel. The capture of this spectral signature enables the architecture to learn discriminative representations of the data. By using the FFA technique, the network aims to update the weights layer by layer in a forward pass only, relieving the computational load of computing the gradient during the backpropagation of the error to update the learned parameters.

The implementation of the 1D CNN layer allows us to implement an FFA architecture similar to the type used for HSI classification \cite{hu2015deep}. The proposed FFA comprises the use of 1D CNN layers in the early stages to capture feature representations within a different latent dimensional space, while the fully connected layers are used at the end to learn how to properly discriminate among the classes. The network is configured as follows: an initial 1D convolutional layer with 64 kernels of size 64, followed by a set of two hidden layers with 128 and 256 feature maps, respectively, both with a kernel size of 36. Then, a max-pooling operation is applied to downscale the dimensions of the tensor by a factor of two. Another 1D CNN layer is applied with 256 kernels of size 36. This last 1D CNN layer is followed by another max-pooling operation with similar characteristics as the previous one, and a flattening operation. Finally, two fully connected layers are added with 100 and $\textit{N}$ units, where $\textit{N}$ represents the number of classes in the HSI dataset.

\subsection{Combination of FFA with backpropagation}
Given the similarities between the nature of FFA and the training procedure in contrasting learning, we propose the utilization of FFA during the initial stage of training. In this stage, each sample is contrasted with different output choices, enabling the model to learn how to effectively discriminate between the correct prediction and other alternatives. Subsequently, the model proceeds to refine its learning through traditional backpropagation using the same deep learning architecture. This process facilitates the model in adjusting the extraction of meaningful characteristics from the high-level spectral information and fine-tuning the latent representation for accurate final predictions.
After this initial phase, the model transitions to the standard backpropagation technique, which leverages the deep learning model's architecture for further refinement. During this stage, the model fine-tunes its representation, optimizing its capacity to capture significant features from the high-dimensional spectral data.

\section{Datasets}\label{sec: Datasets}
For simplicity of this proof-of-concept, we perform experiments over two publicly available dataset\footnote{www.ehu.eus/ccwintco/index.php/Hyperspectral Remote Sensing Scenes}: The Salinas valley and the Indian Pines.
\subsection{The Salinas Valley}
The Salinas dataset is a popular hyperspectral image dataset that is commonly used in the realms of remote sensing and image processing. It is named after the Salinas Valley in California, USA, where the data was collected. The dataset consists of a hyperspectral image of size 512 $\times$ 217 pixels, with 224 spectral bands covering the range from 0.2 to 2.4 micrometers. Each pixel in the image represents a small area on the ground, and the spectral bands capture the reflectance of the surface at different wavelengths. The Salinas dataset was collected using an Airborne Visible/Infrared Imaging Spectrometer (AVIRIS) sensor, which was flown over an agricultural area in the Salinas Valley. The image contains 16 different crop types, including lettuce, broccoli, and bare soil, among others.

\subsection{The Indian Pines}
Collected by the Airborne Visible/Infrared Imaging Spectrometer (AVIRIS) sensor over an agricultural area in Indiana, USA, the Indian Pines dataset consists of a hyperspectral image of size 145 $\times$ 145 pixels, with 224 spectral bands covering the range from 0.4 to 2.5 micrometers. Each pixel in the image represents a small area on the ground, and the spectral bands capture the reflectance of the surface at different wavelengths. The dataset contains 16 different land cover classes, including crops, trees, roads, and buildings, among others. 

\section{Results}\label{sec: results}
\subsection{Experimental setup}

To ensure a fair comparison, we evaluate the different techniques reported in this work using the same training and test datasets. The Salinas and Indian Pines HSI datasets were split into training, validation, and testing sets in an 8:1:1 ratio, respectively. To obtain robust performance estimates, we repeat the entire experimental process three times and average the results.

For each experiment, the respective models are trained for 250 epochs using the Adam optimizer with a fixed learning rate of $1\times10^{-3}$. When employing backpropagation, categorical cross-entropy was utilized as the loss function. However, when FFA is used, a custom loss function is implemented to measure the distance between the \textit{goodness} of each positive and negative sample with respect to the provided threshold. 

All the experiments are run on an NVIDIA RTX 3070 graphic card with 8GB of dedicated GPU. 


\subsection{Performance comparison}
To assess and contrast the performance of the different techniques presented in this work, we employed the following evaluation metrics:
\begin{itemize}
     \item \textbf{Overall Accuracy (OA):} This metric provides the percentage of correctly classified pixels from the respective HSI dataset.
     \item \textbf{Average Accuracy (AA):} This metric is computed by averaging each-class accuracy score, thus providing a class-specific evaluation of the technique's performance. 
     \item \textbf{Kappa Coefficient ($\kappa$):} This metric measures the agreement between the predicted and true class labels, in which the accuracy that could be achieved by chance is taken into account.
\end{itemize}

The use of these evaluation metrics collectively enables a comprehensive assessment of the various techniques presented in this work for HSI classification.

\begin{figure}[ht]%
\centering
\includegraphics[width=1\textwidth]{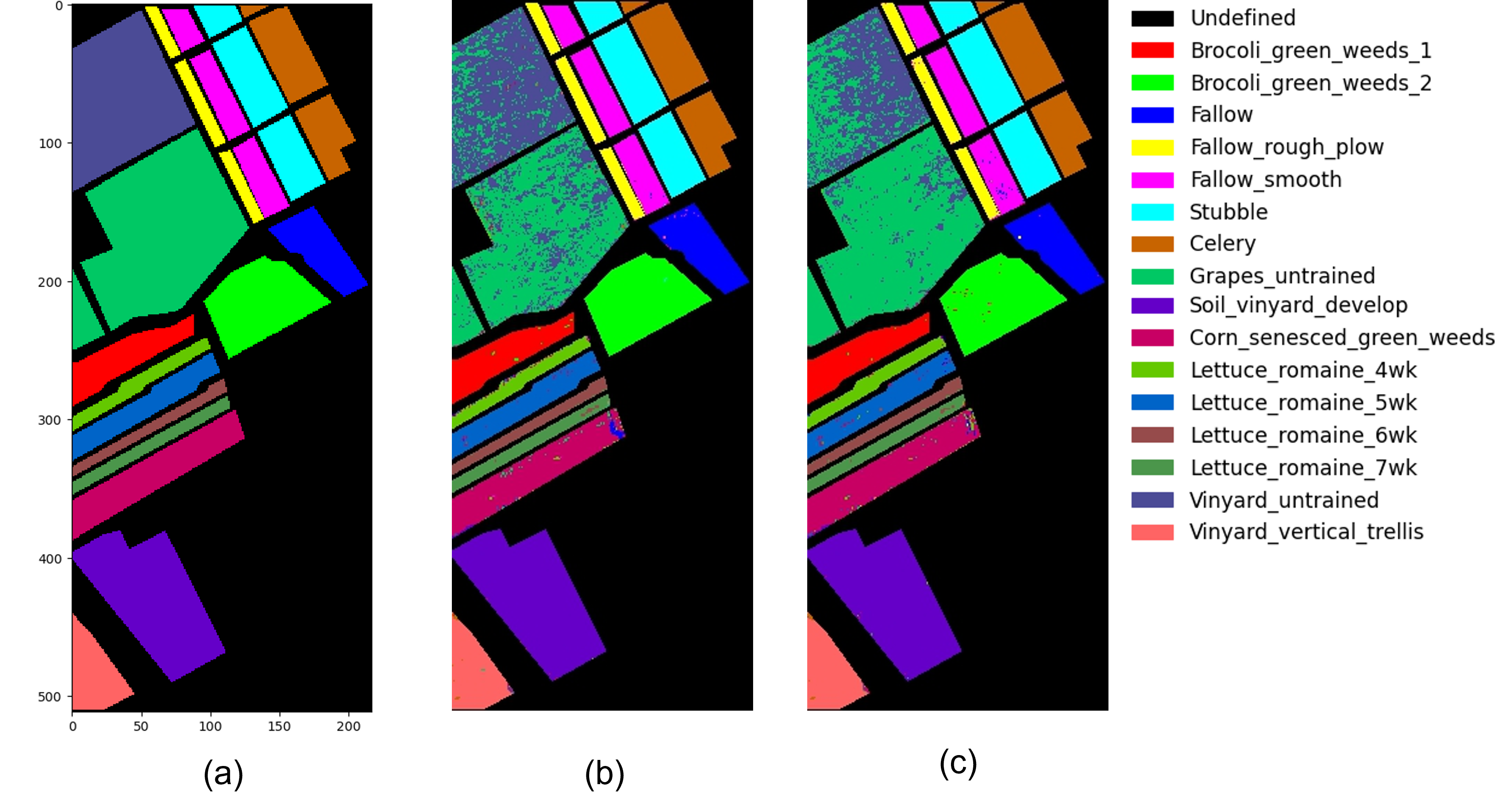}
\caption{Illustration of the predicted classification maps over the Salinas HSI dataset. The images contain (a) the ground truth, (b) the prediction from the use of traditional BP, and (c) the predictions using FFA+BP (c)}\label{fig2}
\end{figure}

\begin{figure}[h!]%
\centering
\includegraphics[width=1\textwidth]{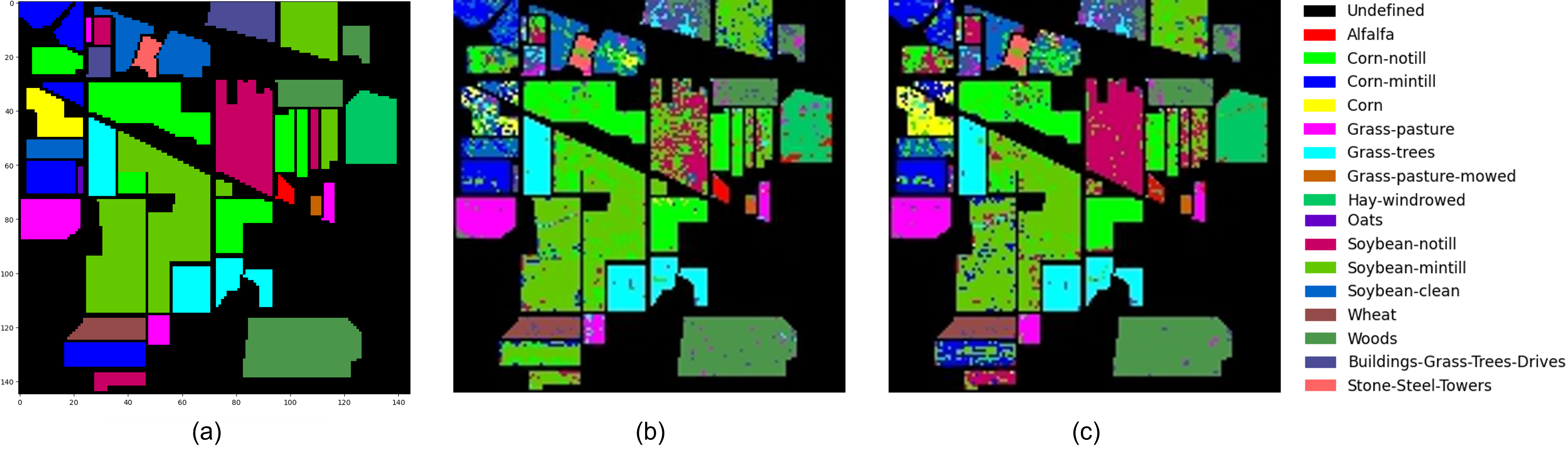}
\caption{Illustration of the predicted classification maps over the Indian Pines HSI dataset. The images contain (a) the ground truth, (b) the prediction from the use of traditional BP, and (c) the predictions using FFA+BP}\label{fig3}
\end{figure}

Table \ref{tab:tab1} summarizes the experimental results obtained by evaluating the discussed techniques using the aforementioned performance metrics on the Salinas and Indian Pines HSI datasets. As shown in Table \ref{tab:tab1}, when considering the Salinas HSI dataset, the combination of FFA and backpropagation (BP) achieved the best performance in terms of OA (0.9221) and $\kappa$ (0.9130). However, BP alone achieved the highest AA (0.9605). On the other hand, for the Indian Pines HSI dataset, BP exhibits the best performance across all evaluation metrics, with OA (0.8109), AA (0.7759), and $\kappa$ (0.7842). Nevertheless, the utilization of FFA in combination with BP demonstrated a significant improvement compared to using any of the FFN variants individually. Figure \ref{fig2} and Figure \ref{fig3} demonstrate classification maps of different models using the Salinas and Indian Pines datasets, respectively.

\begin{table}[]
\caption{Summary of the qualitative results using the different techniques, mentioned in this work, over Salinas and the Indian Pines dataset. Best performance in marked in \textbf{bold} font.}
\label{tab:tab1}
\centering
\scriptsize 
\begin{tabular}{|l|lll|lll|}
\hline
\multicolumn{1}{|c|}{\multirow{2}{*}{\textbf{Method}}} &
  \multicolumn{3}{c|}{\textbf{Salinas}} &
  \multicolumn{3}{c|}{\textbf{Indian   Pines}} \\ \cline{2-7} 
\multicolumn{1}{|c|}{} &
  \multicolumn{1}{c|}{\textbf{OA}} &
  \multicolumn{1}{c|}{\textbf{AA}} &
  \multicolumn{1}{c|}{\textbf{$\kappa$}} &
  \multicolumn{1}{c|}{\textbf{OA}} &
  \multicolumn{1}{c|}{\textbf{AA}} &
  \multicolumn{1}{c|}{\textbf{$\kappa$}} \\ \hline
BP &
  \multicolumn{1}{l|}{0.9190} &
  \multicolumn{1}{l|}{\textbf{0.9605}} &
  0.9099 &
  \multicolumn{1}{l|}{\textbf{0.8109}} &
  \multicolumn{1}{l|}{\textbf{0.7759}} &
  \textbf{0.7842} \\ \hline
FFA (Dense   layers) &
  \multicolumn{1}{l|}{0.8392} &
  \multicolumn{1}{l|}{0.8470} &
   0.8206&
  \multicolumn{1}{l|}{0.5921} &
  \multicolumn{1}{l|}{0.5841} &
   0.5370\\ \hline
FFA (Dense \& Conv layers) &
  \multicolumn{1}{l|}{0.9101} &
  \multicolumn{1}{l|}{0.8518} &
  0.8666 &
  \multicolumn{1}{l|}{0.6660} &
  \multicolumn{1}{l|}{0.5822} &
  0.6194 \\ \hline
FFA + BP &
  \multicolumn{1}{l|}{\textbf{0.9221}} &
  \multicolumn{1}{l|}{0.9564} &
  \textbf{0.9130} &
  \multicolumn{1}{l|}{0.7365} &
  \multicolumn{1}{l|}{0.6978} &
  0.6978 \\ \hline
\end{tabular}
\end{table}

\subsection{Discussion}

In our experiments, we conduct a performance comparison of three different architectures: \textbf{(1)} a deep learning model trained solely using backpropagation, \textbf{(2)} a model trained exclusively with FFA, and \textbf{(3)} our proposed approach that combines FFA pre-training with subsequent fine-tuning using backpropagation. The results clearly illustrate the effectiveness of the combined approach in enhancing feature representation and improving classification accuracy, compared to FFA-only methods.

By combining FFA pre-training with subsequent fine-tuning using backpropagation, we leverage the respective strengths of both approaches. The FFA initializes the network with meaningful representations, which is then fine-tuned through backpropagation to adapt the network to the specific classification task at hand. This combination allows the network to capture both useful spectral characteristics obtained from the FFA and task-specific discriminative features during the backpropagation process. The synergy between these two approaches yields promising results in our experiments.

\section{Conclusion} \label{sec: conclusion}

In summary, the integration of the Forward-Forward algorithm and traditional backpropagation during the early stages of training proved to be highly effective in enhancing feature representation and improving classification performance in hyperspectral image analysis. By incorporating the FFA algorithm, the network was able to capture useful feature representation by adjusting network parameters in every hidden layer. Subsequent fine-tuning through backpropagation facilitated the extraction of discriminative task-specific features. This combined approach exemplifies the potential for leveraging the strengths of different learning algorithms to achieve superior results in hyperspectral image classification tasks.

\bibliographystyle{ieeetr}
\bibliography{ref}

\end{document}